\documentclass[runningheads]{llncs}

\usepackage{eccv}

\usepackage{eccvabbrv}

\usepackage{graphicx}
\usepackage{booktabs}

\usepackage[accsupp]{axessibility}  

\usepackage{hyperref}

\usepackage{orcidlink}

\usepackage{wrapfig} 
\usepackage{graphicx}

\usepackage{amsmath}
\usepackage{amssymb}
\usepackage{algorithm} 
\usepackage{algorithmic} 
\usepackage{enumerate} 
\usepackage{multirow} 
\usepackage{amsmath}

\usepackage{colortbl} 
\usepackage{xcolor}
\usepackage{array}
\usepackage{bbding}
\usepackage{enumerate}
\usepackage{bm}
\usepackage{svg}

\usepackage{helvet} 
\usepackage{courier} 
\usepackage{color}
\usepackage[capitalize,noabbrev]{cleveref}
\usepackage[textsize=tiny]{todonotes}
\usepackage{dsfont}

\newcommand{\Imat}[0]{\ensuremath{{\bf I}} }

\newcommand{\Wmat}[0]{\ensuremath{{\bf W}} }

\newcommand{\cv}[0]{\ensuremath{\boldsymbol{c}} }

\newcommand{\hv}[0]{\ensuremath{\boldsymbol{h}} }

\newcommand{\ov}[0]{\ensuremath{\boldsymbol{o}} }

\newcommand{\qv}[0]{\ensuremath{\boldsymbol{q}} }

\newcommand{\tv}[0]{\ensuremath{\boldsymbol{t}} }

\newcommand{\vv}[0]{\ensuremath{\boldsymbol{v}} }
\newcommand{\wv}[0]{\ensuremath{\boldsymbol{w}} }
\newcommand{\xv}[0]{\ensuremath{\boldsymbol{x}} }
\newcommand{\yv}[0]{\ensuremath{\boldsymbol{y}} }

\newcommand{\Iv}[0]{\ensuremath{\boldsymbol{I}} }

\newcommand{\epsilonv}[0]{\ensuremath{\boldsymbol{\epsilon}} }

\begin{document}

\title{Instruction Tuning-free Visual Token Complement for Multimodal LLMs} 


\author{Dongsheng Wang\inst{1}\orcidlink{0000-0002-3380-5337} \and
Jiequan Cui\inst{2}\orcidlink{0000-0003-2422-7851} \and
Miaoge Li \inst{3}\orcidlink{0009-0006-9531-3254} \and
Wang Lin \inst{4}\orcidlink{0000-0001-8353-2392} \and \\
Bo Chen\thanks{Corresponding author: bchen@mail.xidian.edu.cn} \inst{5} \orcidlink{0000-0001-5151-9388} \and
Hanwang Zhang \inst{2} \orcidlink{0000-0001-7374-8739}}

\authorrunning{D.~Wang et al.}

\institute{College of Computer Science and Software Engineering, Shenzhen University\\
\email{wds\_dana@163.com}
\and
School of Computer Science and Engineering, Nanyang Technological University \and
Department of Computing, Hong Kong Polytechnic University \and
College of Computer Science and Technology, Zhejiang University \and
National Key Lab of Radar Signal Processing, Xidian University}

\maketitle


%
%

\begin{abstract}
As the open community of large language models (LLMs) matures, multimodal LLMs (MLLMs) have promised an elegant bridge between vision and language. However, current research is inherently constrained by challenges such as the need for high-quality instruction pairs and the loss of visual information in image-to-text training objectives. To this end, we propose a Visual Token Complement framework (VTC) that helps MLLMs regain the missing visual features and thus improve response accuracy. Specifically, our VTC integrates text-to-image generation as a guide to identifying the text-irrelevant features, and a visual selector is then developed to generate complementary visual tokens to enrich the original visual input. Moreover, an iterative strategy is further designed to extract more visual information by iteratively using the visual selector without any additional training. Notably, the training pipeline requires no additional image-text pairs, resulting in a desired instruction tuning-free property. Both qualitative and quantitative experiments demonstrate the superiority and efficiency of our VTC.

\end{abstract}

\section{Introduction}
\label{sec:intro}




Large language models (LLMs) have demonstrated extraordinary reasoning ability across various applications~\cite{chatgpt,wei2022chain,touvron2023llama,suris2023vipergpt,zhang2022opt,anil2023palm}. This sparks a research trend that focuses on equipping LLMs with visual prompts to enhance vision-language understanding. To complete various multimodal tasks, we can feed unified vision-language prompts, such as ``\textit{<visual prompt> An image that shows}'' or ``\textit{<visual prompt> What color is the dog?}'', into an LLM to generate the required answers such as image captioning or VQA~\cite{zhu2023minigpt,blip2,liu2023llava,instructblip}. We designate these multimodal-enhanced Large Language Models as Multimodal LLMs (MLLMs), as our focus is solely on multimodal understanding, rather than on both understanding and/or generation. In particular, the latter is known as Large Multimodal Models (LMMs)~\cite{alayrac2022flamingo,li2023otter,wang2022git,driess2023palm,zhao2023making,chen2022pali}, which, unfortunately, have been shown to possess a lower understanding capability compared to MLLMs~\cite{instructblip,yin2023survey,fu2023mme}.

\begin{figure}[!t]
\centering
\includegraphics[width=0.84\linewidth]{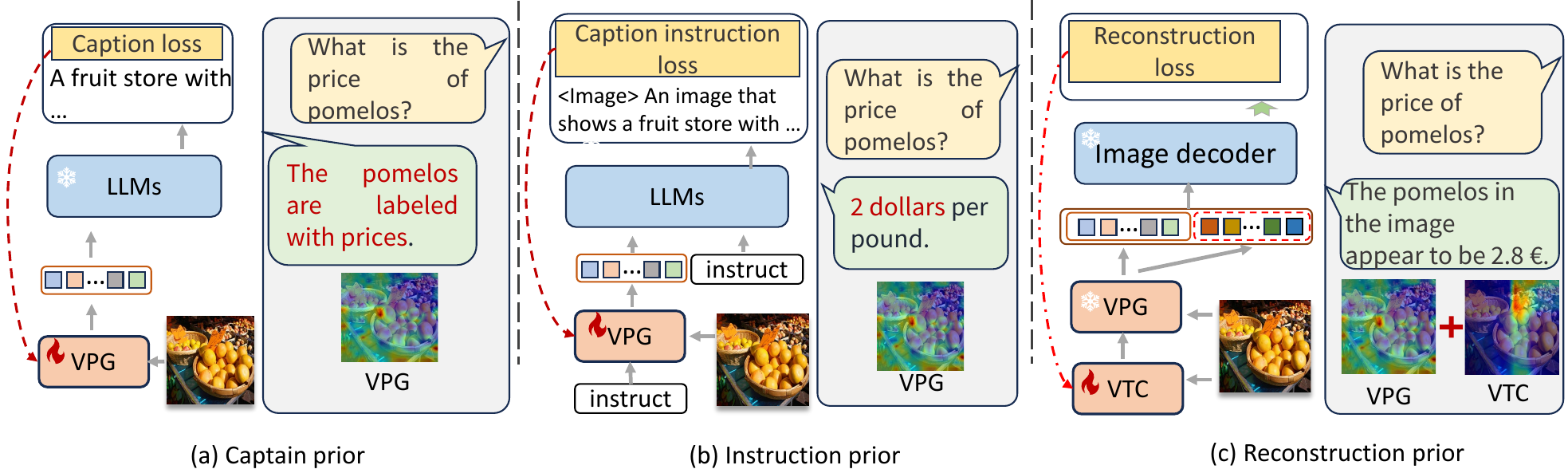}
\caption{\small{Comparisons of our proposed Reconstruction prior (c) for training Visual Token Complement (VTC), Caption prior (a), and Instruction prior (b) for training Visual Prompt Generator (VPG). The attention maps of VPG and VTC are obtained by averaging the query-patch attention weights(Sec.~\ref{sec:vtc}).}}
\label{vpgs_com}
\end{figure}





    


As ``an image is more than a thousand words'', it is efficient to represent visual prompts as continuous embeddings but not discrete language tokens. For example, BLIP-2~\cite{blip2} and MiniGPT4~\cite{zhu2023minigpt} use Q-Former to transform image patches into a fixed number of visual tokens; while LLaVA~\cite{liu2023llava} demonstrates that a simple linear projection can be equally effective instead of using Q-Former. These visual tokens are expected to encode the visual contents in the image, providing informative context for LLMs to generate responses.


The crux of making  MLLMs effective is to train a Vision Prompt Generator (VPG) that can generate visual tokens, recognizable by LLMs, from an input image. As VPG is essentially a visual-to-semantic translator, we need a visual-semantic alignment training task to align the visual tokens and their semantics written in language output by a frozen LLM. There are two representative tasks that introduce different visual priors into VPG:

\noindent\textbf{Caption prior}: Image captioning is a straightforward alignment task. For instance, the Q-Former VPG in BLIP-2~\cite{blip2} and MiniGPT4~\cite{zhu2023minigpt} is trained on large-scale image-caption pairs~\cite{laion400m}: the frozen LLM should output the ground-truth caption when taking the visual tokens of the corresponding image as input. However, this task inevitably introduces an inductive bias losing significant details, that is, the VPG tends to only capture the visual semantics which is just enough for captioning the image contents according to the prescribed annotation details in training. As illustrated in Fig.~\ref{vpgs_com}(a), as the VPG only focuses on the main object \textit{pomelos} but neglects the \emph{price tag}, the MLLM fails to answer the question about price. 

\noindent\textbf{Instruction prior}: Inspired by instruction tuning~\cite{wei2021finetuned,ouyang2022training,zhang2023instruction},  which turns the general language prior of LLMs---next-token prediction---into more powerful task solvers, InstructBLIP~\cite{instructblip} and LLaVA~\cite{liu2023llava} aim to train VPG for task-aware visual tokens. They curate a variety of vision-language tasks such as image caption, visual grounding and VQA into the same <image, instruction, output> template, which is fed into a frozen LLM for training VPG. However, curating instructions at scale and with diversity is particularly costly for vision-language tasks, so the VPG is usually biased to the instruction prior due to overfitted tuning. As shown in Fig.~\ref{vpgs_com}(b), the VPG cannot distinguish the \emph{price tags} of different fruits because such price inquiry of multiple objects is rare in the multi-task datasets used in InstructBLIP~\cite{instructblip}.


\begin{figure}[!t]
\centering
\includegraphics[width=0.82\linewidth]{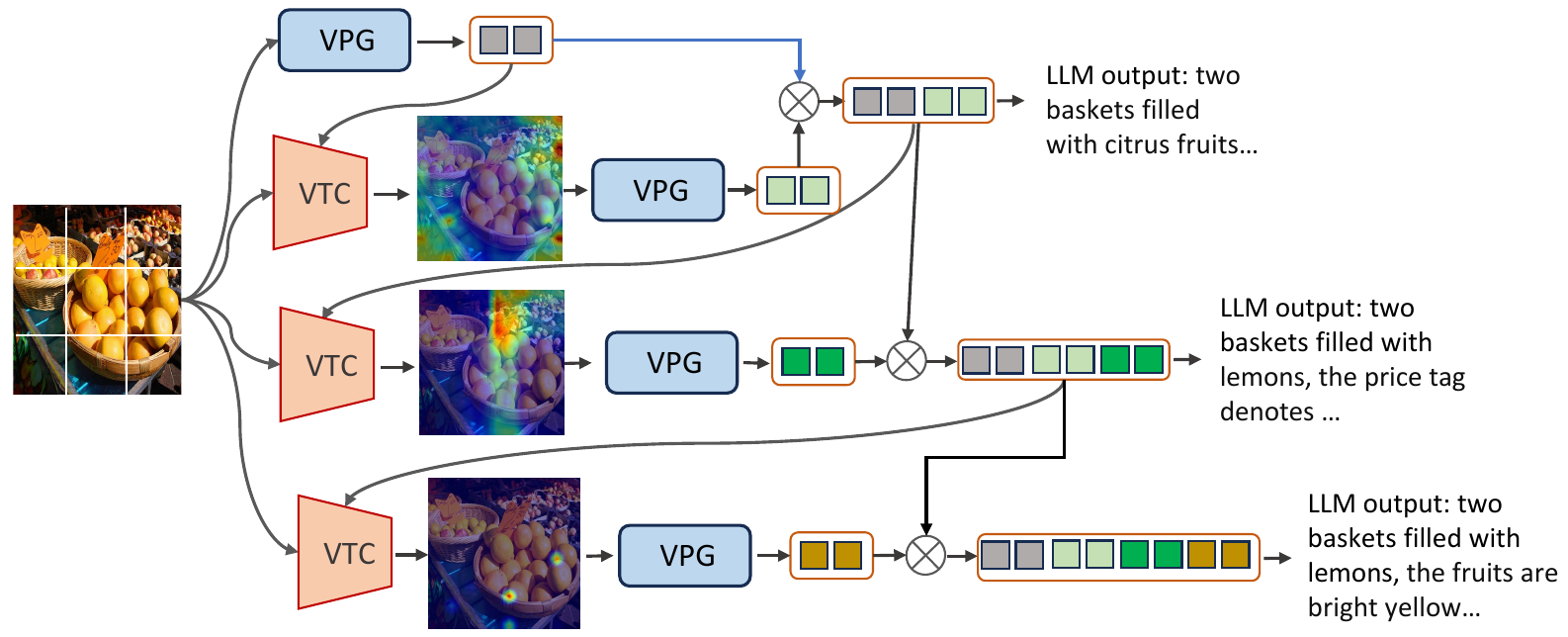}
\caption{\small{Iterative complement at the inference stage. Once trained, Our VTC can be reused multiple times to re-pick up the visual concepts omitted by the previous VPG. We showcase three times complements with increasingly detailed output captions.}}
\label{fig:2}
\end{figure}

In this paper, we use \textbf{Reconstruction prior} to recover the missing visual semantics by VPG with caption/instruction prior. As shown in Fig.~\ref{vpgs_com}(c), we propose a trainable module called \textbf{Visual Token Complement (VTC)} that attends to visual regions missed by VPG. By feeding them to VPG, we can obtain the complementary visual tokens encoding visual semantics not yet captured by VPG. Together with the existing visual tokens, how well the complementary tokens complete the image contents is measured by the image reconstruction loss. We highlight the following three advantages of the proposed VTC:
\begin{itemize}
\item\textbf{Instruction tuning-free}. Training VTC is unsupervised. All we need to calculate the reconstruction loss is the input images and three frozen modules: VPG, LLM, and image decoder. No caption pairs are needed to train our VTC. 

\item\textbf{Completeness on-demand}. As illustrated in Fig.~\ref{fig:2}, thanks to the simple implementation of VTC (Sec.~\ref{sec:vtc}), we have the flexibility to iteratively generate additional complementary tokens as required. As the VPG is frozen, the iteration can always make up for the visual semantics missed by the VPG in the previous reconstruction. We show that two iterations are sufficient in our experiments (Sec.~\ref{analysis}).

\item\textbf{Vision-language disentanglement}. Thanks to the above advantages,  if we have instruction tuning data, we can fix the trained VTC and VPG and only tune the LLM for instruction comprehension. As the visual tokens are already completed, such disentanglement can mitigate the vision-language spurious correlation in training data, and utilize the valuable instruction data more efficiently---LLM is more specialized in understanding instructions. We show promising results in Sec.~\ref{analysis}.
\end{itemize}

To evaluate the proposed VTC, we conducted extensive comparisons on \textit{LVLM-eHub}~\cite{xu2023lvlm}, \textit{MME}~\cite{fu2023mme}, and \textit{DEMON}~\cite{li2023finetuning} multimodal benchmarks, covering 22 diverse vision tasks in the zero-shot setting. By completing the missing information, our VTC shows significant improvements over its baselines, especially in visual dialogue (showing 45\% improvement over InstructBLIP~\cite{instructblip}). Besides, the visualizations of the appended visual tokens, provide an interpretable tool to analyze the conplementary semantics.

\section{Related Work}
Recently, research focused on extending large language models (LLMs) to accommodate image inputs has garnered increasing attention, primarily due to their significant improvements in various vision-language tasks~\cite{wu2023visual,chatgpt4v,gao2023llama,li2023finetuning}. They aim to learn cross-modal reasoning from large-scale multimodal data. 
Pioneering studies like SimVLM~\cite{wang2021simvlm}, Frozen~\cite{tsimpoukelli2021multimodal}, and GiT~\cite{wang2022git} view image as a foreign language and have achieved exceptional fine-tuning performance under the autoregressive framework. In-context learning series~\cite{alayrac2022flamingo,li2023otter,awadalla2023openflamingo,zhao2023mmicl} are proposed to align a pre-trained vision encoder with LLMs by the design of multimodal prompt, showing remarkable potential in zero-shot and few-shot tasks. Following that, various VPGs are developed to efficiently align visual features with the language model, \textit{e.g.}, Linear mapping~\cite{liu2023llava} and Q-Former~\cite{blip2}. The follow-up instruction-tuned methods are proposed from various assumptions to help VPGs make full use of prior knowledge implied in LLMs. For instance, LLaVA~\cite{liu2023llava} collects high-quality instruction pairs and finetunes the linear projection layer between the VPG and LLM. InstructBLIP~\cite{instructblip} introduces instruction-aware Q-Former and extracts informative features according to the given instruction. Several attempts have also been made to enable MLLMs to understand videos and audios. They share similar techniques with image-language alignment, yielding impressive performance. The scope of this paper is image-language modalities.

However, most existing VPGs align the vision and language domains through autoregressive training objectives, \textit{e.g.} the image-to-caption loss, which may result in incomplete alignment due to the semantic gaps and noising matchings in multimodal data~\cite{zhao2023making}. Our approach, in contrast,  adopts a text-to-image strategy to guide the learning of visual tokens so that they can reconstruct the input image as closely as possible, resulting in instruction tuning-free visual token complement.

\section{The Proposed Method}
For a given image $\xv$, VPGs aim to generate a soft vision prompt $\yv =\{\yv_i\}_{i=1}^N$ containing $N$ visual tokens, where $\yv_i \in \mathbb{R}^d$ is the $n$-th token vector and $d$ is the embedding dimension.
Without loss of generality, we adopt the Q-Former as the VPG in the following discussion. Q-Former utilizes a fixed number of $N$ query vectors $\qv$ to interact with image features $\hv=f(\xv)$ by multiple cross-attention layers, where $f$ is a frozen vision encoder (\textit{e.g.}, ViT-G). The output queries are used as visual tokens $\yv$, which are prepended in the input text embedding of the instruction. To train a VPG, an autoregressive loss is often employed on the predicted text: $p(\wv|\yv,\wv_{\text{instruct}})$, where $\wv_{\text{instruct}}$ represents the textual input and $\wv$ is the predicted answers. Caption prior-based VPGs typically specify $\wv_{\text{instruct}}$ with the caption template: ``<visual prompt> An image that shows''; while the instruction prior-based VPGs tailor various templates to specific tasks, \textit{e.g.}, ``<visual prompt> Question: \{Question\}. Short answer:'' for VQA and ``<visual prompt> Given the image, generate a question whose answer is: \{Answer\}. Question:'' for visual grounding. VPGs serve as an intermediary between the image and the frozen LLM, with their core idea being to find the optimal $\yv$ that provides efficient visual semantics about $\xv$, thereby enabling the LLM to produce accurate responses.


\begin{figure*}[!t]
\centering
\includegraphics[width=0.88\linewidth]{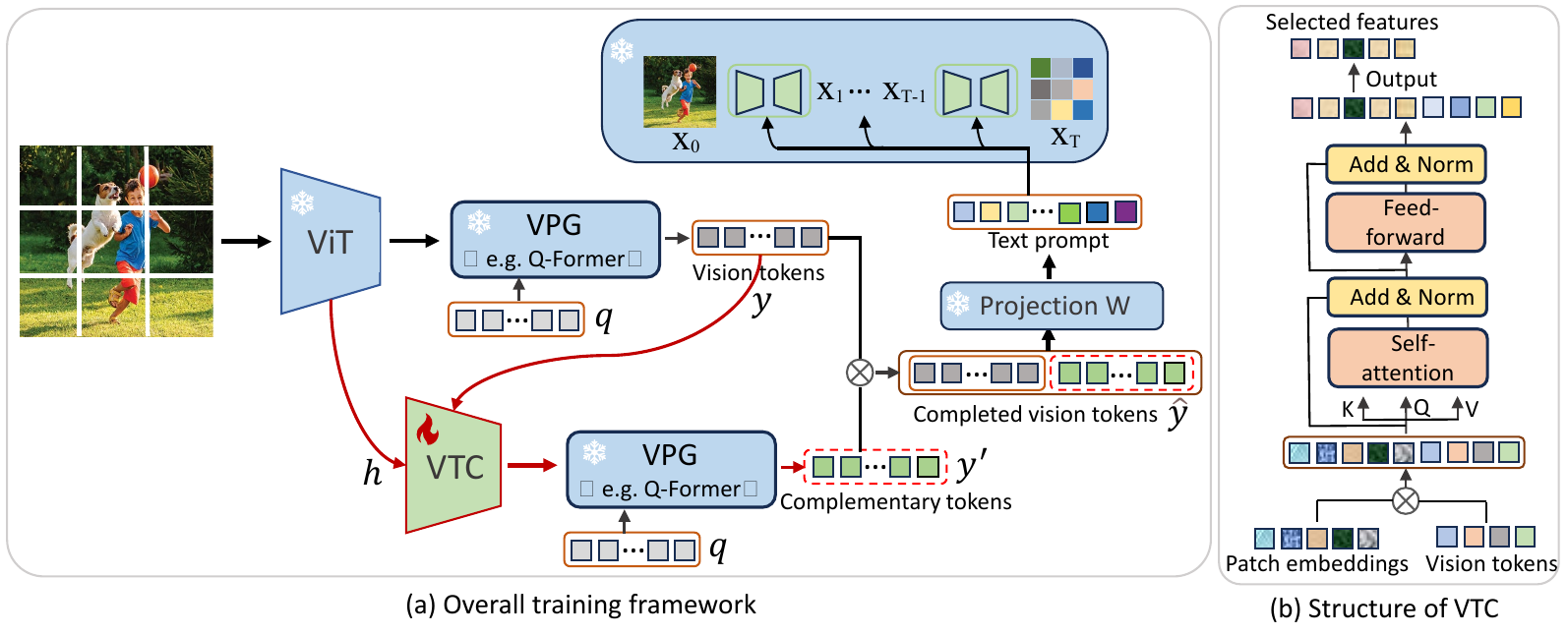}
\caption{\small{(a) Training pipeline of the proposed VTC, and (b) structure of VTC. }}
\label{framework}
\end{figure*}


\subsection{Visual Token Complement} \label{sec:vtc}
In this paper, we focus on addressing the issue of information loss during transforming image features into visual tokens, and propose the Visual Token Complement (VTC) framework for existing VPGs. As discussed above, VPGs based on caption/instruction prior often struggle with incompleteness issues, where the generated vision tokens $\yv$ are prone to emphasizing main objects at the expense of finer image details. To complete $\yv$ and provide comprehensive visual details into LLM, VTC introduces an additional set of vision tokens $\yv' =\{\yv_i'\}_{i=1}^N$ by reusing the VPG:
\begin{equation} \label{selector}
    \yv' = \text{VPG}(\cv, \qv), \quad \cv=\text{VTC}(\hv, \yv),
\end{equation}
where VTC takes image feature $\hv$, the generated vision tokens $\yv$ as inputs and outputs the complementary feature $\cv$. We will introduce the detailed structure of VTC later. $\cv$ captures the visual details missed by $\yv$ and is fed into the VPG to output the complementary visual tokens. Mathematically, our VTC can be applied with most existing VPGs, \textit{e.g.}, Q-Former and linear projection in LLaVA~\cite{liu2023llava}\footnote{We empirically adopt the Q-Former as our VPG. LLaVA generates 256 tokens for each image, which is too long for training our VTC.}



Once the complementary vision tokens $\yv'$ are generated, the final completed vision tokens are obtained as:
\begin{equation}
    \hat{\yv} = \text{Concat}(\yv, \yv'),
\end{equation}
where $\text{Concat} (\cdot)$ denotes applying the concatenating at the token dimension. Like previous MLLMs, $\hat{\yv}$ in VTC will be inserted into the position of the virtual image tokens to form the LLM's input according to different instruction templates. Crucially, the completed $\hat{\yv}$ combines the original caption-aware $\yv$ and the newly generated tokens $\yv'$, where the former tends to capture the main visual objects and the latter focuses on the missing details. Together, they offer comprehensive visual evidence and help LLM capture more vision semantics.

In addition, moving beyond the patch embeddings at the last layer of $f$, VTC aims to explore multi-level semantics and provide hierarchical vision descriptions for LLMs. 
Let $\hv^{(l)}$ denote the patch embeddings at $l$-th layer of $f$, VTC pickups the missing features as:
\begin{equation} \label{selector}
    [{\cv}^{(l)},\ov] = \text{Self-Att}(\text{Concat}(\hv^{(l)}, \yv)), 
\end{equation}
where $\text{Self-Att}$ denotes a standard Self-Attention layer, which stacks the self-attention, feeding forward, and residual layers (shown as Fig.~\ref{framework}(b)). $\cv^{(l)}$ denotes the complementary feature at layer $l$. Naturally, we can obtain the multi-layer complementary features $\hv$ in Eq.~\ref{selector} by concatenating $\cv^{(l)}$ at different layers, resulting in more comprehensive visual tokens $\hat{\yv}$.



\subsection{Reconstruction-prior Training}
Inspired by autoencoders~\cite{kenton2019bert, he2022masked} that perform representation learning by reconstructing the input, we here introduce the reconstruction-prior training to learn the optimal $\hat{\yv}$. Specifically, we view $\hat{\yv}$ as the latent embeddings and learn VTC with the image reconstruction task. To complete the encoder-decoder framework, we employ a frozen stable diffusion (SD) ~\cite{rombach2022high, song2021scorebased} as the image decoder $g(\xv|\hat{\yv})$, mapping $\hat{\yv}$ back to the pixel space due to its impressive text-to-image results in the recent AI-generated content field~\cite{saharia2022photorealistic,khachatryan2023text2video,zhang2023text}. 

Given $(\xv, \hat{\yv}) $, SD views $\hat{\yv}$ as the textual condition, and updates it by solving the following Gaussian noise regression as in training the reversed diffusion steps (with $T$ timesteps):
\begin{equation}\label{loss}
    \arg \min_{\hat{\yv}} \mathbb{E}_{\xv_0, \epsilonv, t} [||\epsilonv - \epsilon(\xv_t, \hat{\yv} \Wmat)||^2],
\end{equation}
where $\xv_0=\xv$, $\epsilonv \sim \mathcal{N}(0,\Iv)$ is the Gaussian noise, and $t\sim [0,T]$ is the sampled timestep. $\epsilon (\cdot)$ is the frozen noise prediction network, conditionally on the completed $\hat{\yv}$, $\xv_t$ is noised image at timestep $t$: $\xv_t \sim \mathcal{N}(\sqrt{1-\beta_t}\xv_{t-1},\beta_t \Iv)$ with $\beta_t \in (0,1)$ being the variance. 
$\Wmat \in \mathbb{R}^{d \times d_c}$ is a linear mapping that projects the $d$-dimensional LLM embedding into the $d_c$-dimensional CLIP space. We pre-calculate $\Wmat$ via the least squares with orthogonality constraint~\cite{paischer2023sitta}:
\begin{equation} \label{pre_w}
\begin{aligned}
    \arg \min_{\Wmat} \mathbb{E}_{v \in V_{\text{CLIP}} \cap V_{\text{LLM}}} &||\vv_{\text{CLIP}} - \vv_{\text{LLM}}\Wmat||^2, \\
   & \textit{s.t.}, \Wmat^T \Wmat = \Imat,
\end{aligned}
\end{equation}
where $V_{\text{CLIP}}$ and $V_{\text{LLM}}$ denote the vocabulary sets of CLIP and LLM, respectively, and $\vv_{\text{CLIP}}$ and $\vv_{\text{LLM}}$ are the corresponding CLIP and LLM embedding vector of the shared token $v$. The orthogonality constraint ensures the learned projection matrix $\Wmat$ implicitly preserves the structure of the LLM's embedding and regulars the learning of $\Wmat$ focus on the semantic translation\cite{schonemann1966generalized,paischer2023sitta}.

Unlike previous VPGs that optimize the vision tokens $\yv$ by minimizing the image-to-text loss, – with $\yv$ inserted into the ``<vision prompt>''part of the instruction.
Eq.~\ref{loss} gives us an elegant solution to learn the optimal $\hat{\yv}$ based on the reconstruction prior. The idea is simple: the vision tokens are complete if they can reconstruct the input image. Leveraging the reconstruction loss, the proposed VTC is trained in an unsupervised manner. Given the frozen image encoder, VPG and SD, VTC is the only to-be-learned module to regain the missing features for better reconstruction. This disentangles the vision-language learning. 
Since the visual tokens $\hat{\yv}$ are already completed, we can fine-tune the LLM for instruction comprehension on instruction pairs, which is the same as the pre-training stage of LLM, where LLM is more specialized.

\subsection{Iterative Inference}
Once trained with Eq.~\ref{loss}, our VTC in Eq.~\ref{selector} has the ability to identify the missing features by comparing the generated tokens $\yv$ and the image embedding $\hv$. This makes it possible to reuse VTC and then iteratively generate additional complementary tokens as required. Specifically, let $\yv^k$ and ${\hat{\yv}}^k$ denote the generated vision tokens and the completed vision tokens at iteration $k$, respectively, \textit{e.g.}, $\yv^1=\yv'$ and ${\hat{\yv}}^1=\hat{\yv}$. Thus the complementary tokens ${\yv}^{k+1}$ can be obtained as:
\begin{equation}
    {{\yv}}^{k+1} = \text{VPG}(\cv^k, \qv), \quad \cv^k = \text{VTC}(\hv, {\hat{\yv}}^k).
\end{equation}
$\yv^{k+1}$ captures the visual details missed by ${\hat{\yv}}^k$. Benefiting from the flexible structure of VTC, we can train VTC at $k=1$ and iteratively reuse it to squeeze the visual message.
Empirically, we find two iterations are sufficient, accounting for the computational cost.

\section{Experiments}
We conduct extensive experiments on several benchmark multimodal datasets to evaluate the zero-shot performance of VTC against the state-of-the-art MLLMs.
\subsection{Experimental Settings}
\noindent\textbf{Three multimodal benchmarks}.
In the experiment, we evaluate the diverse and robust performance of our proposed VTC on three zero-shot multimodal benchmarks, \textit{LVLM-eHub}~\cite{xu2023lvlm}, \textit{MME}~\cite{fu2023mme} and \textit{DEMON}~\cite{li2023finetuning}. They measure the visual-language comprehension ability of MLLM from different perspectives. \textit{LVLM-eHub} is the first multimodal benchmark, which focuses on visual knowledge acquisition, visual reasoning, embodied intelligence, and other 6 categories of multimodal capabilities. \textit{MME} is another recently proposed benchmark that measures both the visual perception and cognition abilities of MLLMs across 14 datasets. \textit{DEMON} focuses on more complex scenarios, where the demonstrative instructions may contain multiple, interleaved, and multimodal inputs. It covers 29 tasks across 7 categories, including multimodal dialogue, visual storytelling, and other multimodal understanding scenarios.


\noindent\textbf{Baselines}. We compare our VTC with the recent advanced MLLMs on three widely used benchmarks. For all methods, we choose versions with parameter sizes less than 10B, Otter~\cite{li2023otter}, which aligns the image and text under the in-context learning framework; mPLUG-Owl~\cite{ye2023mplug} tunes the vision encoder and LLM (with a LoRA module~\cite{hu2021lora}) at different stages; VPGTrans~\cite{zhang2023transfer}, which designs a projector warm-up and vanilla fine-tuning to transfer an existing VPG from one MLLM to the target MLLM; And VPG based MLLMs: LLaVA~\cite{liu2023llava}, BLIP2~\cite{blip2}, MiniGPT-4~\cite{zhu2023minigpt}, and InstructBLIP~\cite{zhang2023instruction}, they share similar ideas but varying VPGs and training data. For all methods, we select versions with parameter sizes under 10B. Please refer to the Appendix for details. 

\noindent\textbf{Training Data}. We train our VTC in an unsupervised manner, and we adopt the ImageNet1K as our training dataset.

\noindent\textbf{Implementation Details}. We chose three VPG-based MLLMs as our base model: BLIP2, MiniGPT4, and InstructBLIP, and we denote the corresponding variants as VTC(BLIP2), VTC(MiniGPT4), and VTC(InstructBLIP). We follow the same implementation details as the base model. For the SD model, we load the checkpoints of Stable-Diffusion-v1-5. VTC is trained on ImageNet1K with 2 epochs. For the learning rate, we first conduct a linear warm-up from $1e-6$ to $1e-4$, and then use a cosine learning rate schedule with the minimal $lr=1e-5$. We set $l=[12,24,36]$ in Eq.~\ref{selector} to capture the multi-level visual semantics.

\subsection{Quantitative Analysis}
\textbf{Results on \textit{LVLM-eHub}}~\cite{xu2023lvlm}. We first evaluate the vision-language understanding of our VTC on \textit{LVLM-eHub}. We here focus on the Visual Reasoning task due to it requires a comprehensive understanding of images and texts. It includes visual question answering (VQA, \textit{e.g.},DocVQA, TextVQA, STVQA, OCR-VQA,
OKVQA, GQA, and IconQA), visual spatial reasoning (VSR), visual dialog (\textit{e.g.}, Visdial), and knowledge-grounded image description (\textit{e.g.}, ScienceQA and VizWiz). We report the evaluation results at Table.~\ref{lvlm}. We have the following remarks about the results. Overall, our proposed VTC outperforms the others consistently on all datasets, except the case where ours is the second on the DocVQA dataset. Moreover, compared with our base models (BLIP2, MiniGPT4, and InstructBLIP), our model performs better in general, especially in visual dialog cases. This is not surprising, as our VTC acts as a complementary module and provides LLM with useful image details, which are usually missed by VPGs. By concatenating the original tokens with the complementary tokens and freezing the whole MLLM, VTC improves VPGs by making full use of the prior knowledge.

\begin{table*}[t]
    \centering
    \scalebox{0.78}{
   \begin{tabular}{l|cccccccccccc}
   \toprule[1.5pt]
   Methods &Doc &Text &STV &OCR &OK &GQA &Visdial &Icon &VSR &WHOOPS &ScienceQA &VizWiz \\
   \midrule
   mPLUG-Owl &2.24 &38.76 &12.10 &8.84 &22.89 &14.02 &13.34 &11.64 &24.74 &20.70 &2.80 &38.99 \\
   Otter &3.44 &21.52 &15.23 &19.50 &49.01 &38.12 &11.67 &26.77 &6.40 &15.14 &27.22 &50.04\\
   VPGTrans &3.53 &21.98 &17.13 &21.71 &44.51 &32.99 &9.70 &38.22 &48.77 &15.88 &36.99 &53.23\\
   \hline
    LLaVA &\textbf{6.26} &38.92 &28.40 &23.40 &54.36 &41.30 &14.66 &42.95 &51.24 &24.39 &49.33 &62.42\\
   BLIP2 &4.75 &31.98 &20.98 &38.85 &44.93 &45.53 &10.73 &60.82 &63.63 &24.87 &60.73 &65.44 \\
   \rowcolor{gray!40} VTC(BLIP2) &5.32 &32.0 &25.67 &38.89 &46.08 &48.97 &\textbf{77.20} &\textbf{62.31} &\textbf{63.77} &26.65 &\textbf{63.10} &{66.30} \\
   MiniGPT-4 &2.65 &19.40 &13.55 &16.85 &37.48 &30.82 &10.31 &37.59 &41.56 &17.91 &25.43 &47.48 \\
   \rowcolor{gray!40} VTC (MiniGPT-4) &3.04 &22.75 &15.50 &19.54 &41.75 &35.11 &76.62 &39.03 &43.75 &18.53 &37.85 &52.10 \\
   \hline
   InstructBLIP &5.89 &{39.60} &28.30 &60.20 &60.52 &49.96 &45.20 &56.25 &41.28 &30.13 &46.26 &65.31 \\
   \rowcolor{gray!40}VTC (Instruct) &{6.14} &\textbf{40.52} &\textbf{30.70} &\textbf{60.24} &\textbf{60.96} &\textbf{51.12} &65.78 &56.84 &43.10 &\textbf{33.74} &46.21 &\textbf{66.31} \\
   \bottomrule[1.5pt]
    \end{tabular}}
    \caption{\small{Comparison of Zero-shot Performance for various MLLMs on general VQA tasks.}}
    \label{lvlm}
\end{table*}


\begin{table*}[!th]
    \centering
    \scalebox{0.73}{
   \begin{tabular}{l|ccccc>{\columncolor{gray!40}}cc>{\columncolor{gray!40}}cc>{\columncolor{gray!40}}c}
   \toprule[1.5pt]
   Datasets &mPLUG-Owl &Otter &VPGTrans &LLaVA &BLIP2 &VTC(B) &MiniGPT4 &VTC(M) &InstructBLIP &VTC(I) \\
   \midrule
  Existence &120.00 &\textbf{195.00} &70.00 &50.00 &160.00 &170.00 &68.33 &72.00 &185.00 &188.00\\
  Count &50.00 &88.33 &85.00 &50.00 &135.00 &138.39 &55.00 &55.00 &143.33 &\textbf{144.01}\\
  Position &50.00 &\textbf{86.67} &63.33 &50.00 &77.33 &80.51 &43.33 &45.76 &66.67 &70.84\\
  Color &55.00 &113.33 &73.33 &55.00 &148.33 &152.90 &75.00 &75.00 &153.33 &\textbf{154.51}\\
  Poster &136.06  &138.78 &84.01 &50.00 &141.85 &\textbf{147.32} &41.84 &44.00 &123.81 &127.04\\
  Celebrity &100.29 &\textbf{172.65} &53.53 &48.82 &105.59 &112.65 &54.41 &57.63 &101.18 &108.74\\
  Scene &135.50 &\textbf{158.75} &141.75 &50.00 &145.25 &150.00 &71.75 &75.34 &153.00 &155.00\\
  Landmark &\textbf{159.25} &137.25 &64.75 &50.00 &138.00 &141.28 &54.00 &54.00 &79.75 &80.34\\
  Artwork &96.25 &129.00 &77.25 &49.00 &136.50 &\textbf{137.44} &60.50 &64.39 &134.25 &134.97\\
  OCR &65.00 &72.50 &77.50 &50.00 &110.00 &\textbf{110.00} &57.50 &60.00 &72.50 &77.50\\
  \hline 
  Commonsense &78.57 &106.43 &64.29 &57.14 &110.00 &110.95 &59.29 &62.55 &129.29 &\textbf{130.03}\\
  Numerical &60.00 &\textbf{72.50} &50.00 &50.00 &40.00 &42.00 &45.00 &45.00 &40.00 &42.00\\
  Text Translation &\textbf{80.00} &57.50 &77.50 &57.50 &65.00 &65.00 &5.50 &60.00 &65.00 &67.50\\
  Code Reasoning &57.50 &70.00 &57.50 &50.00 &\textbf{75.00} &70.50 &40.00 &40.00 &57.50 &60.00\\
   \bottomrule[1.5pt]
    \end{tabular}}
    \caption{\small{Evaluation results (\%) of perception and recognition on \textit{MME}. The results are calculated by summing ACC and ACC+ in \textit{MME}\cite{fu2023mme}. We shorten VTC variants by keeping the first letter of the corresponding base models.}}
    \label{mme}
\end{table*}

\noindent\textbf{Results on \textit{MME}}~\cite{fu2023mme}. \textit{MME} is another recently proposed benchmark measuring both the visual perception and cognition abilities of MLLMs across 14 datasets. Given a test image, \textit{MME} designs two prompts, and each of them consists of two parts, including a visual question and an instruction ``Please answer yes or no''. This makes it convenient to report the quantitative performance of accuracy. We report the accuracy score at Table.~\ref{mme}. First, We find that Otter, VTC(B), and VTC(I) achieve 5/14, 3/14, and 3/14 best results, respectively. Otter aligns the vision-language modalities under the in-context learning framework on massive multimodal datasets, while our approach is trained from a single modality. These competitive results indicate our unsupervised training paradigms effectively complete the missing semantics, showing great potential in overcoming the inherent limitation of VPGs. Second, compared to mPLUG-Owl (with 2/14 best results), which finetunes the LLM for multimodal alignments, our approach shows better performance in terms of both perception and cognition. This indicates the efficiency of our image reconstruction loss, which helps VTC complete the image details so that the LLM has sufficient visual evidence for the correct response.

\begin{figure}[!t]
    \centering
    \begin{minipage}{0.48\textwidth}
        \centering
        \includegraphics[width=\linewidth]{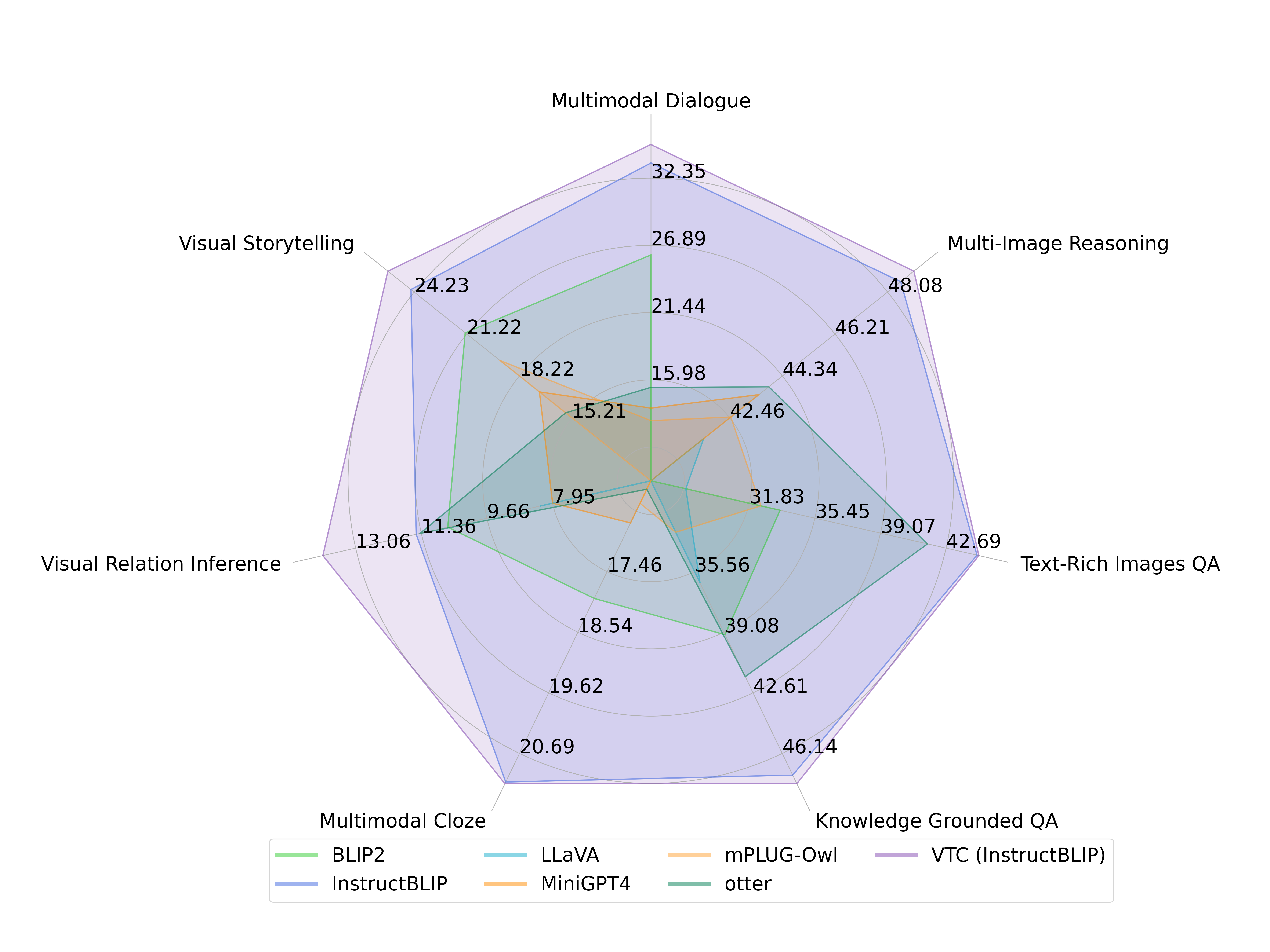} 
        \caption{\small{Zero-shot evaluation results on \textit{DEMON}. We only report our VTC(InstructBLIP) for clear representation.}}
        \label{demon}
    \end{minipage}\hfill 
    \begin{minipage}{0.48\textwidth}
        \centering
        \includegraphics[width=\linewidth]{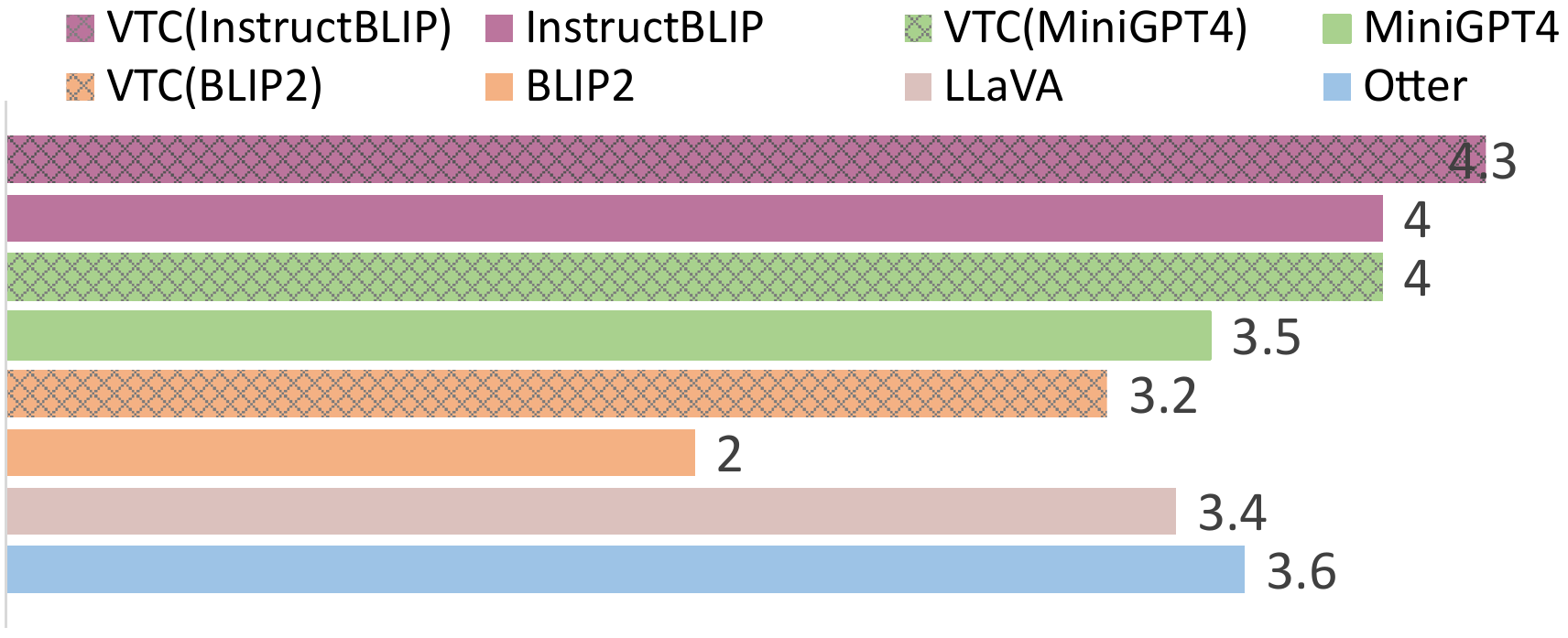} 
        \caption{\small{ChatGPT 4(V) evaluation results of various MLLMs with scores ranging from [1-5].}}
        \label{chatgpt}
    \end{minipage}
\end{figure}

\noindent\textbf{Results on \textit{DEMON}}~\cite{li2023finetuning}. We adopt \textit{DEMON} as our third vision-instruction understanding benchmark. Different from the first two evaluations, \textit{DEMON} designs more complex demonstrative instructions, including interleaved, and multimodal inputs.
We report the evaluation score in Fig.~\ref{demon}. First, our approach surrounds other baseline methods in all directions, demonstrating robust multimodal understanding. Moreover, our approach obtains 20\% improvement over InstructBLIP in visual relation inference. We attribute this to our reconstruction prior, which offers helpful guidance to help visual tokens memorize the location information of the targets. Lastly, compared to Otter, which shows higher performance on \textit{MME}, our VTC(InstructBLIP) achieves significant improvements on all tasks. This demonstrates our superior vision-language understanding in complex scenarios, where the reconstruction prior still provides fine-grained guidance, while the captions fail to capture such detailed supervision.


\subsection{Further Analysis} \label{analysis}

\noindent\textbf{ChatGPT 4(V) evaluation}. Besides the above three benchmark evaluations, we further leverage ChatGPT 4(V)~\cite{chatgpt4v} to have a systematic measurement of the quality of the proposed model. Specifically, motivated by previous works~\cite{vicuna2023, liu2023llava} we first randomly select 50 image-instruction pairs from the MIMIC-IT datasets~\cite{li2023mimicit} and collect the answers from various MLLMs. ChatGPT 4(V) is then asked to evaluate the quality of answers based on the input image and output the corresponding scores (1-5), where a higher score indicates better overall performance. We report the results in Fig.~\ref{chatgpt}. We find that our VTC(InstructBLIP) is the only model with a score larger than $4$, and the second model is our VTC(MiniGPT4). We hope this evaluation results serve as strong evidence for supporting our performance. The evaluation prompt in this section can be found in the Appendix.

\begin{figure*}[!t]
\centering
\includegraphics[width=0.84\linewidth]{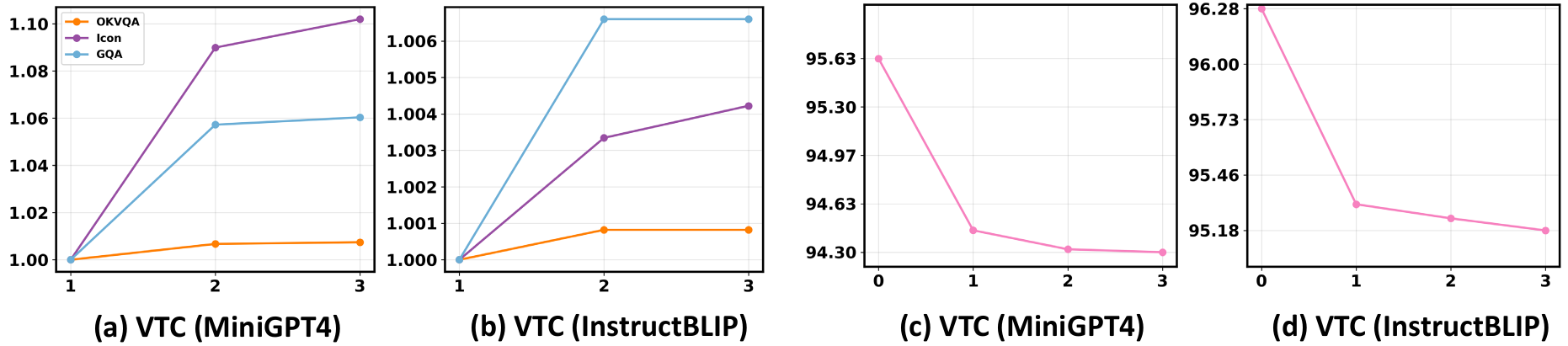}
\caption{\small{(a-b) Improvement trends on OKVQA, IconQA, and GQA datasets with different numbers of iterations. (c-d) tag-token nearest neighbor semantic distance ($\times 100$) on Flicker dataset with different bumbers of iterations ($x=0$ means the original VPG).}}
\label{iteration}
\end{figure*}

\begin{figure*}[!t]
\label{vpgs}
\centering
\includegraphics[width=0.99\linewidth]{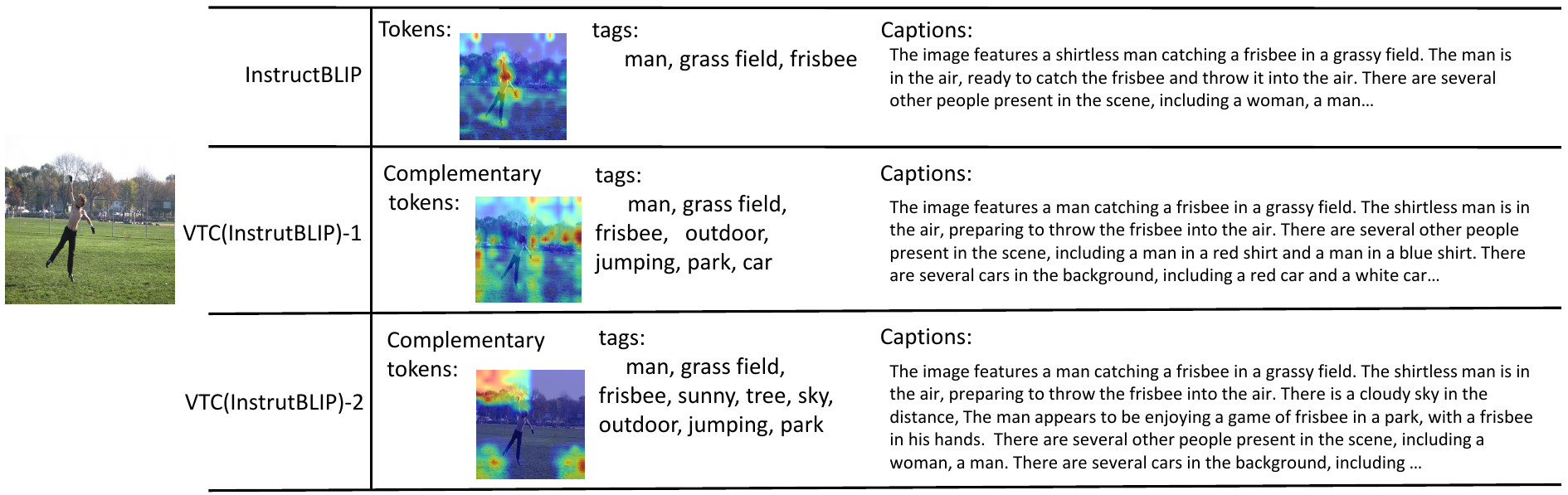}
\caption{\small{Visulization of the complementary tokens in the first iteration (VTC(InstructBLIP)-1) and second iteration (VTC(InstructBLIP)-2). We also provide the retrieved tags and predicted captions.}}
\label{token_vis}
\end{figure*}

\noindent\textbf{Information completion of VTC}. One of the core ideas behind VTC is to learn complementary visual tokens that capture visual features missed by VPG. To evaluate such information completion, we here propose the tag-token distance $d$, which measures the nearest neighbor semantic distance between the image tags and generated visual tokens: $d=\frac{1}{M}\sum_{i=1}^M \text{cosine}(\tv_i, \yv_i)$, where $M$ is the number of tags, the visual token $\yv_i$ is the nearest neighbor of tag $i$, and $\text{cosine} (\cdot)$  is the cosine distance. We adopt the tagged Flicker dataset~\cite{huiskes2008mir} to measure the semantic distance, where the average number of tags per image is 8.94. We report the results of VTC(MiniGPT4) and VTC(InstructBLIP) in Fig.~\ref{iteration}(c-d), where the $x$-axis denotes the number we reuse the VTC module, and $x=0$ means the original VPG. We find that VTC significantly reduces the semantic distance. That is, the newly generated tokens are able to capture visual concepts missed by the previous tokens, which chimes with our motivations. We also visualize the vision tokens in Fig.~\ref{token_vis}, which will be introduced later.

\begin{table}[t]
    \centering
    \scalebox{0.90}{
   \begin{tabular}{l|ccc|ccc}
   \toprule[1.5pt]
   \multirow{2}{*}{Datasets} &\multicolumn{3}{c|}{VTC(BLIP2)} &\multicolumn{3}{c}{VTC(InstructBLIP)} \\
   \cline{2-7}
    ~ &BLIP2 &BLIP2(64) &VTC &InstructBLIP &InstructBLIP(64)  &VTC\\
    \hline
    Perception &133.79 &133.95  &138.23 &121.42 &121.86  &124.31 \\
    Cognition &73.94 &73.96 &75.54 &73.02 &73.44 &75.92 \\
    \bottomrule[1.5pt]
   \end{tabular}}
   \caption{\small{Abalation results of fine-tuning with LoRA and the number of visual tokens. We finetune LLMs in our model and baselines with LoRA to test the performance of finetuning with the instruction dataset. (64) means we add 32 visual queries to the original model, thus resulting visual tokens have the same length as our VTC.}}
   \label{fn}
\end{table}

\noindent\textbf{Impact of the number of iterations}. Due to the flexible design of our VTC, we can reuse the VTC module iteratively at the test stage, once the VTC is trained with the first iteration. This makes it possible to increasingly complete the visual tokens on demand. Yet more iterations mean higher computational costs. We report the impact of the number of iterations in Fig.~\ref{iteration}. Fig.~\ref{iteration}(a-b) denote the normalized VQA results on OKVQA, IconQV, and GQA datasets with different iterations(\textit{e.g.},1,2,3), and Fig.~\ref{iteration}(c-d) denote the tag-token distance on Flick dataset. Overall, increasing the number of iterations can improve the performance in terms of VQA and tag-token retrievals. This demonstrates that the newly generated complementary tokens indeed capture more visual concepts, enhancing the LLMs. Besides, we observe that two iterations are sufficient and after two, the gain is not very significant. This is reasonable since our frozen SD model forces the visual tokens $\hat{\yv}$ to capture as many features as possible to reconstruct the input image in the first iteration, and therefore there is little detail to pick up in the following iterations.

\begin{figure*}[!t]
\label{demos}
\centering
\includegraphics[width=.80\textwidth]{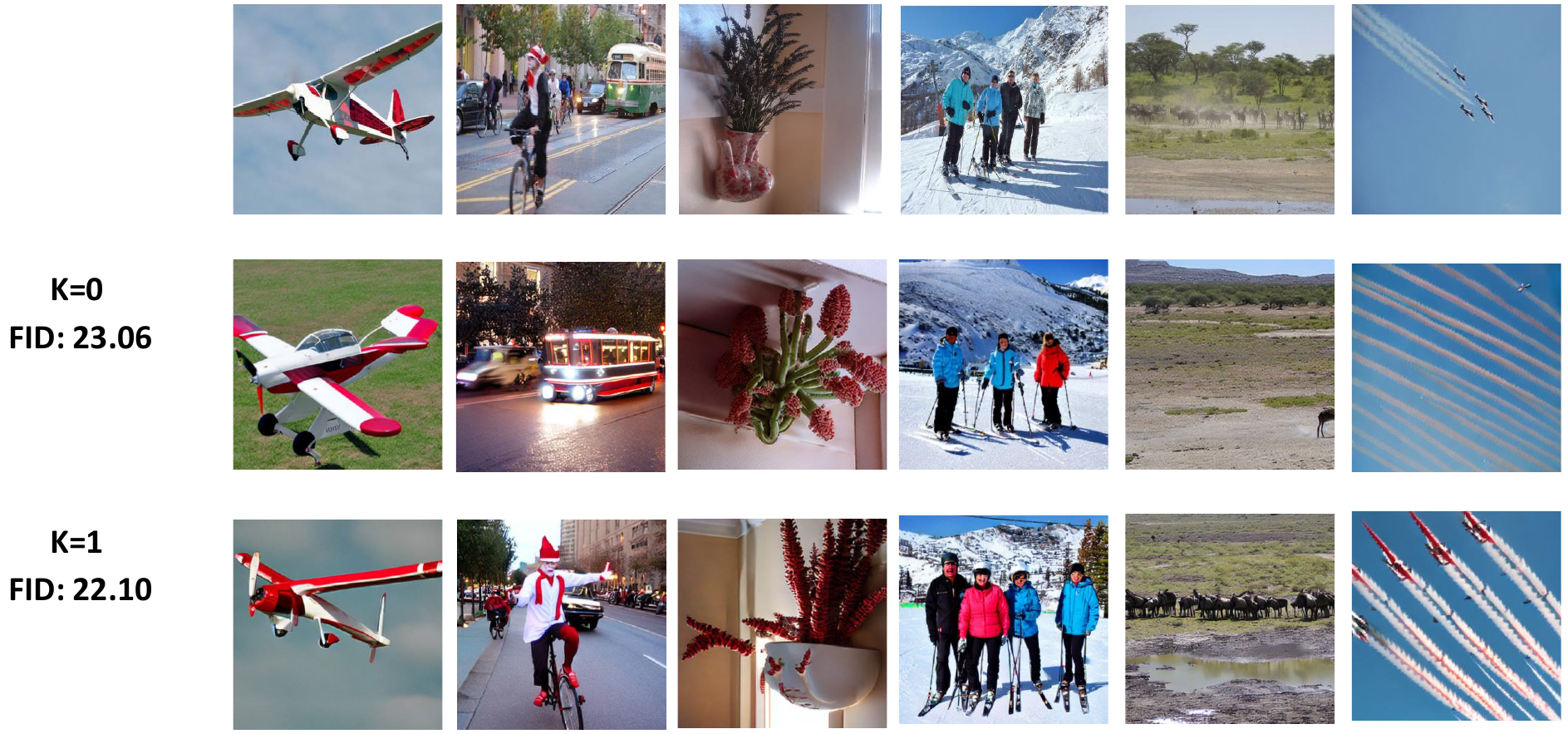}
\caption{\small{Ablation results of the reconstruction quality. The first row denotes the raw input images. $K=0$ denotes the original InstructBLIP and $K=1$ denotes we use VTC once to complete the visual tokens.}}
\label{vis_app}
\end{figure*}

\begin{figure*}[!t]
    \centering
    \begin{subfigure}[b]{\linewidth}
    \centering
        \includegraphics[width=0.8\textwidth]{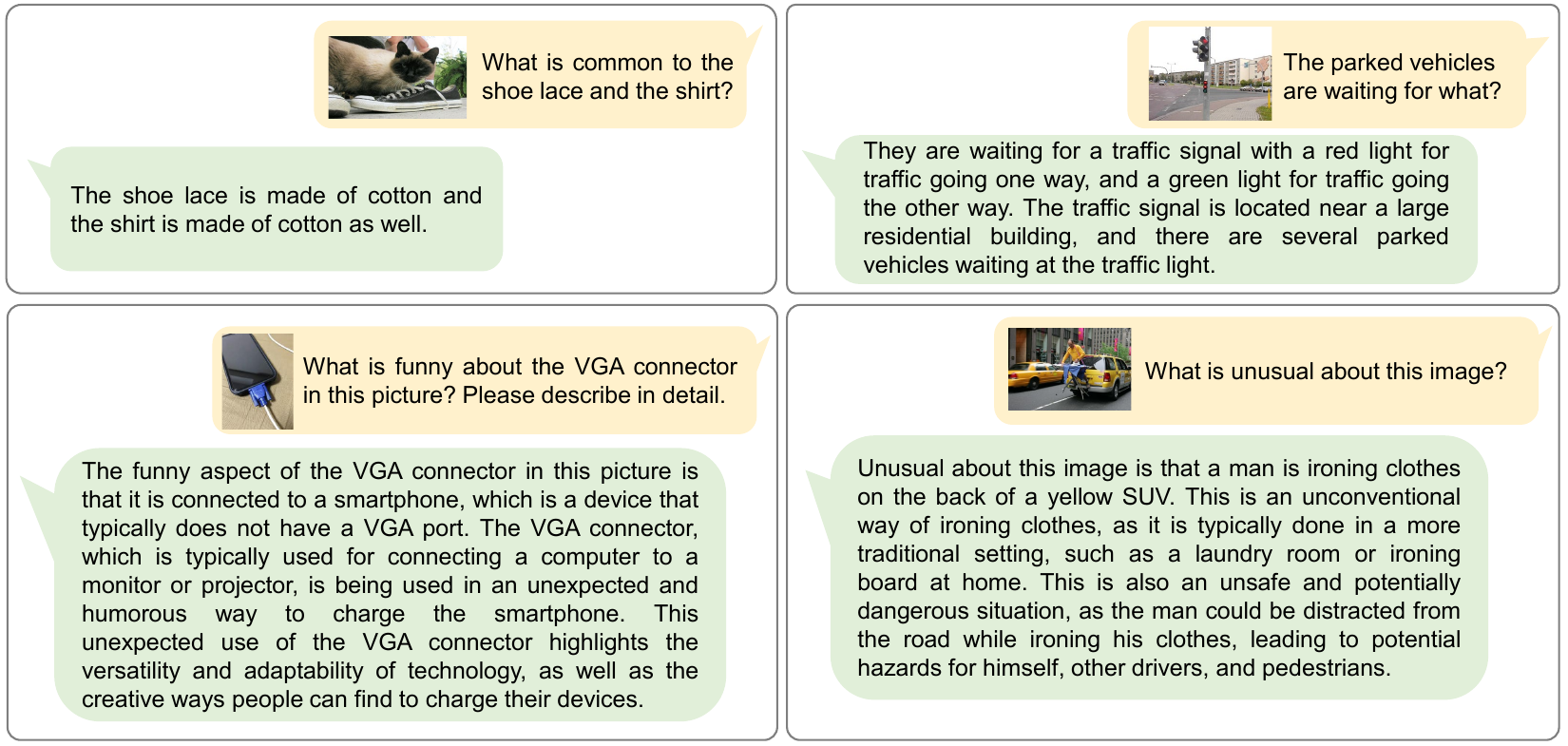}
        \label{fig:image1}
    \end{subfigure}

    \begin{subfigure}[b]{\linewidth}
    \centering
        \includegraphics[width=0.8\textwidth]{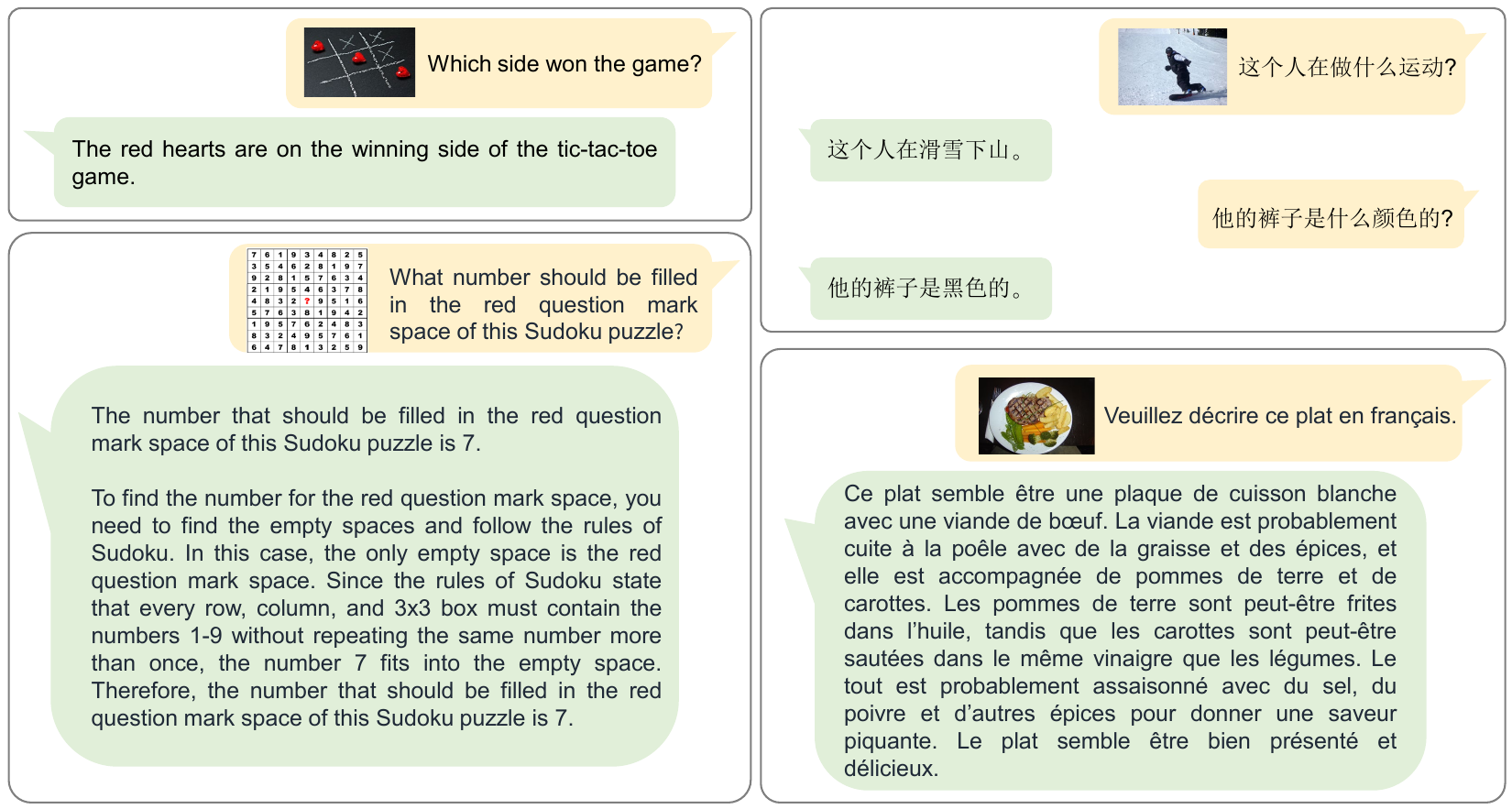}
        \label{fig:image2}
    \end{subfigure}

    \begin{subfigure}[b]{\linewidth}
    \centering
        \includegraphics[width=0.8\textwidth]{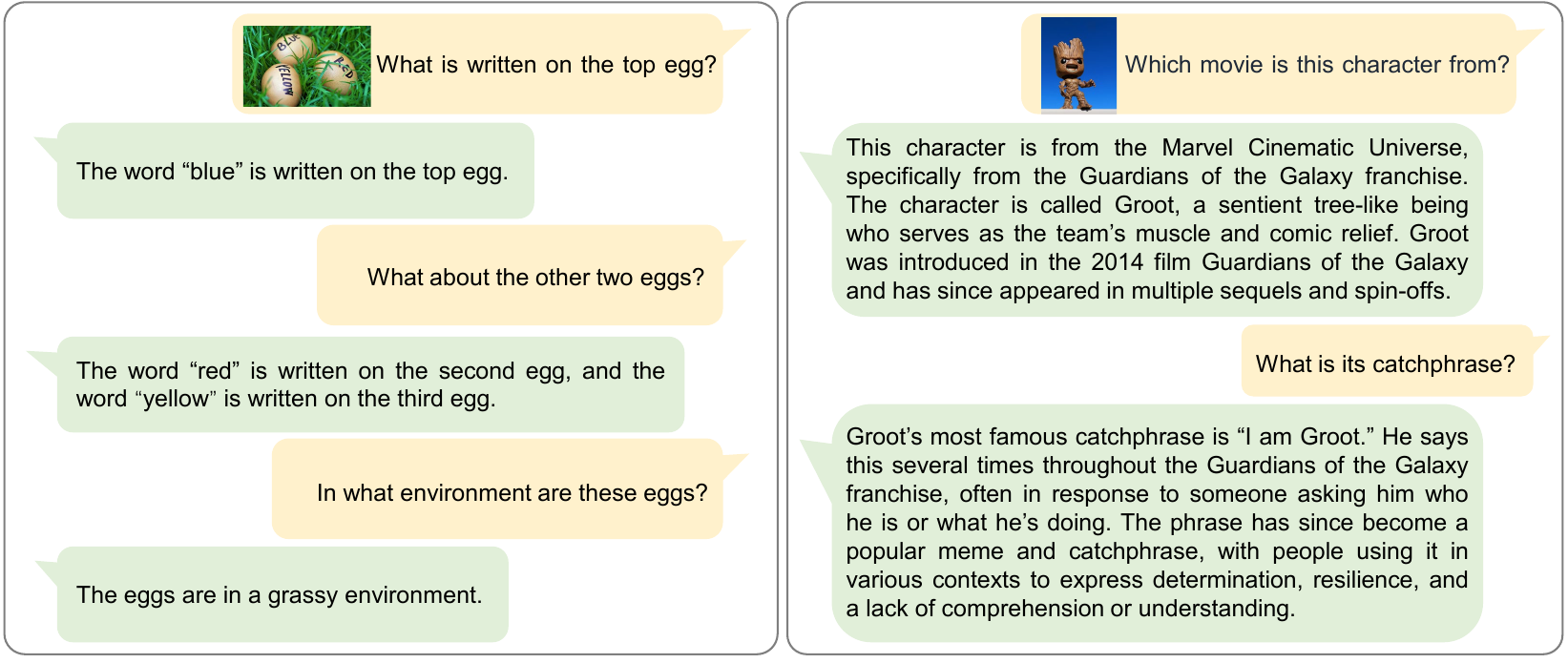}
        \label{fig:image2}
    \end{subfigure}
    
    \caption{\small{Examples of the test dialogs. VTC shows impressive vision-language understanding in image captioning, human common sense, and visual reasoning, across single and multi-turns.}}
    \label{conversation}
\end{figure*}

\noindent\textbf{Vision-language disentanglement}. After being trained with reconstruction prior, the completed tokens $\hat{\yv}$ provide LLMs with lossless image messages. Thus we can fix the VTC and VPG and only finetune the LLM for instruction learning. This mitigates the spurious correlations between vision translating and instruction understanding, which are two important components of MLLMs and often entangled with each other in previous VPGs. As the visual tokens are already completed, we can focus on the latter by finetuning LLMs. Specifically, We finetune our LLM and baselines with LoRA~\cite{hu2021lora} on the LLaVA Visual Instruct 150K dataset~\cite{liu2023llava} and report the evaluation results on \textit{MME} benchmark at Table.~\ref{fn}. We find that fine-tuning LLM improves performance considerably further. Recalling that BLIP2 is trained with image-caption pairs and InstuctBLIP tunes BLIP2 with massive instruction pairs, our VTC(BLIP2) with finetuning outperforms VTC(InstructBLIP) on both the perception and cognition tasks. This demonstrates that our approach uses the instruction pairs more efficiently. Besides, we also test the performance of visual query length. Our VTC learns 32 additional tokens in the visual prompt, we thus add 32 visual queries to the original baselines and jointly train those vectors on the instruction pairs. We find that our VTC achieves higher scores on \textit{MME}. We attribute this to our reconstruction-aware training, which effectively learns missing concepts.

\noindent\textbf{Ablation studies on reconstruction quality}. One of the core ideas behind VTC is to learn reconstruction-aware tokens for comprehensive image representation. The proposed VTC utilizes a frozen SD as the image decoder to complete the visual tokens. To measure the performance of the completed visual tokens, We here aim to test the improvements by measuring the quality of the reconstructed images. Specifically, we report the FID scores on the COCO2014 validation dataset. A lower FID score means closer semantics between the reconstructed images and the original images. From Fig.~\ref{vis_app} we find that VTC captures more image details, such as the number of persons, the number of airplanes, and the accurate sky background. The higher reconstruction quality demonstrates the efficiency of the newly generated tokens, thus contributing to the responses of LLMs.

\subsection{Visualization \& Case Studies}
Recall that the cross-attention weights between the query embeddings $\qv$ and the patch embeddings $\hv$ in VPG provide us with an interpretable tool to visualize the complementary tokens. Specifically, we average the attention weight over all tokens and visualize the attention map in Fig.~\ref{token_vis}. In addition to the attention map, we also provide the retrieved tags and the corresponding predicted caption. We find that the complementary tokens successfully attend to the missing regions and different iterations cover diverse visual concepts. This meets with our complement motivation, resulting in more accurate tags and detailed captions.

We also showcase several test dialogs in Fig.~\ref{conversation}. Those examples cover human common sense, reasoning, and knowledge retrieval, across single and multiple turns. For example, our VTC has the ability to reason with human common sense, identify the funny aspect of connecting VGA with smartphones, and the unusual cases in a street picture. Moreover, VTC is able to infer the winner of a tic-tac-toe game and the correct number in a Sudoku puzzle based on pre-trained human knowledge. Those results demonstrate the impressive visual perception ability of our approach.

\section{Conclusion}
We in this paper propose VTC, an instruction tuning-free framework to learn complementary visual tokens that capture image details missed by VPG. VTC leverages text-to-image generation to guide the acquisition of reconstruction-aware vision tokens. The enhanced vision prompt is formed by merging these new tokens with the original ones. Additionally, we have developed an iterative strategy to progressively reclaim missing visual information on demand during the testing phase Extensive experiments are conducted on three widely used multimodal benchmarks to evaluate the proposed model. A ChatGPT 4(V) evaluation is also included to have a systematic measurement. Visualization and case studies demonstrate the interpretability and impressive vision-language understanding of our model.

\section*{Acknowledgement}
This work was supported in part by the National Natural Science Foundation of China under Grant U21B2006; in part by Shaanxi Youth Innovation Team Project; in part by the Fundamental Research Funds for the Central Universities QTZX24003 and QTZX22160; in part by the 111 Project under Grant B18039.

\bibliographystyle{splncs04}
\bibliography{main}

\clearpage
\appendix
\setcounter{page}{1}
\renewcommand\thetable{\Alph{section}. \arabic{table}}   \renewcommand\thefigure{\Alph{section}. \arabic{figure}}    
\setcounter{table}{0}
\setcounter{figure}{0}
\begin{figure*}[h]
    \centering
    \begin{subfigure}[b]{\linewidth}
        \includegraphics[width=\textwidth]{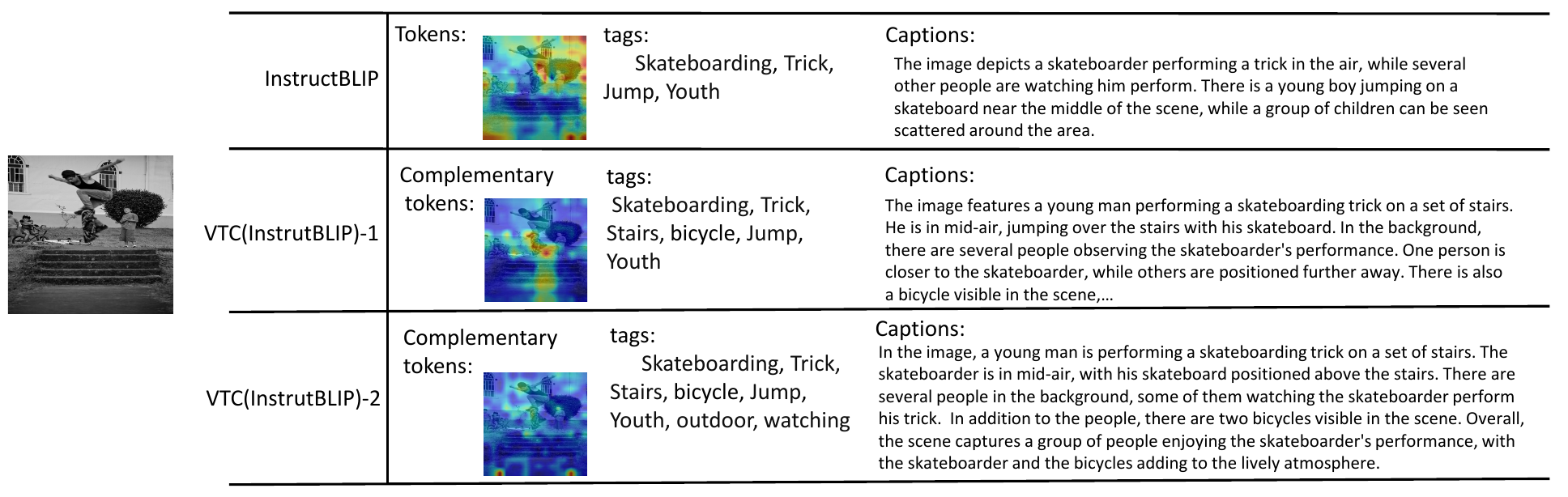}
        \label{fig:image1}
    \end{subfigure}

    \begin{subfigure}[b]{\linewidth}
        \includegraphics[width=\textwidth]{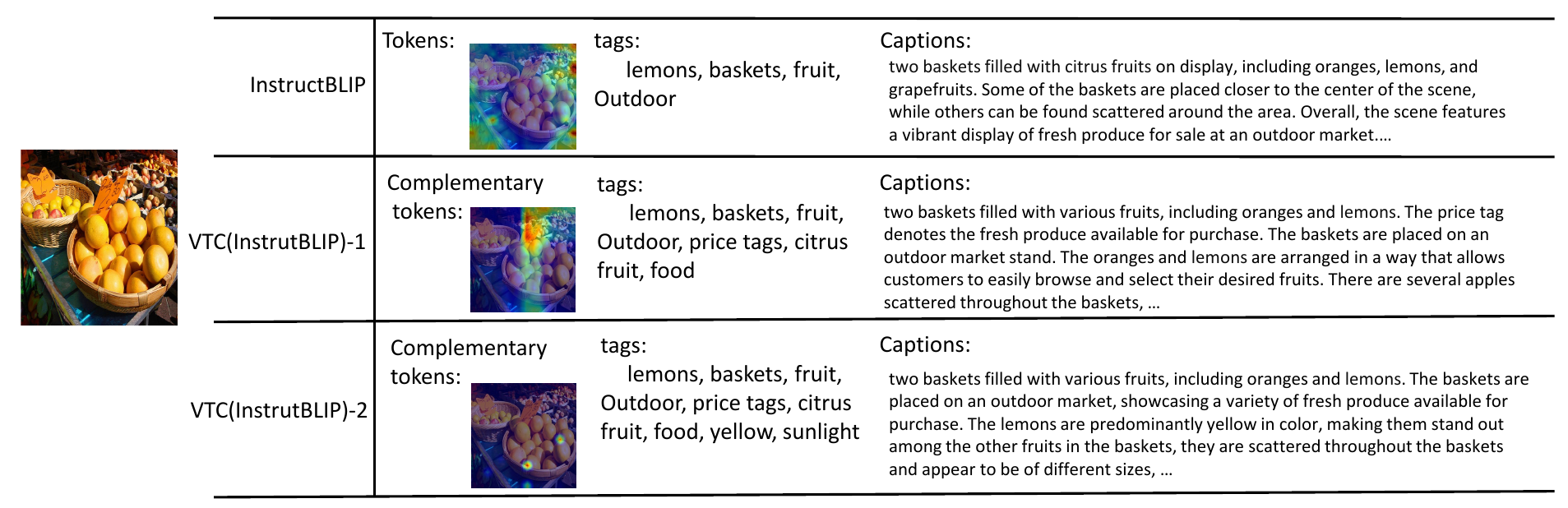}
        \label{fig:image2}
    \end{subfigure}
    
    \caption{\small{More visualizations of the complementary tokens in the first iteration (VTC(InstructBLIP)-1) and second iteration (VTC(InstructBLIP)-2).
We also provide the retrieved tags and predicted captions.}}
    \label{app_park}
\end{figure*}

\begin{figure*}[h]
\centering
\includegraphics[width=\textwidth]{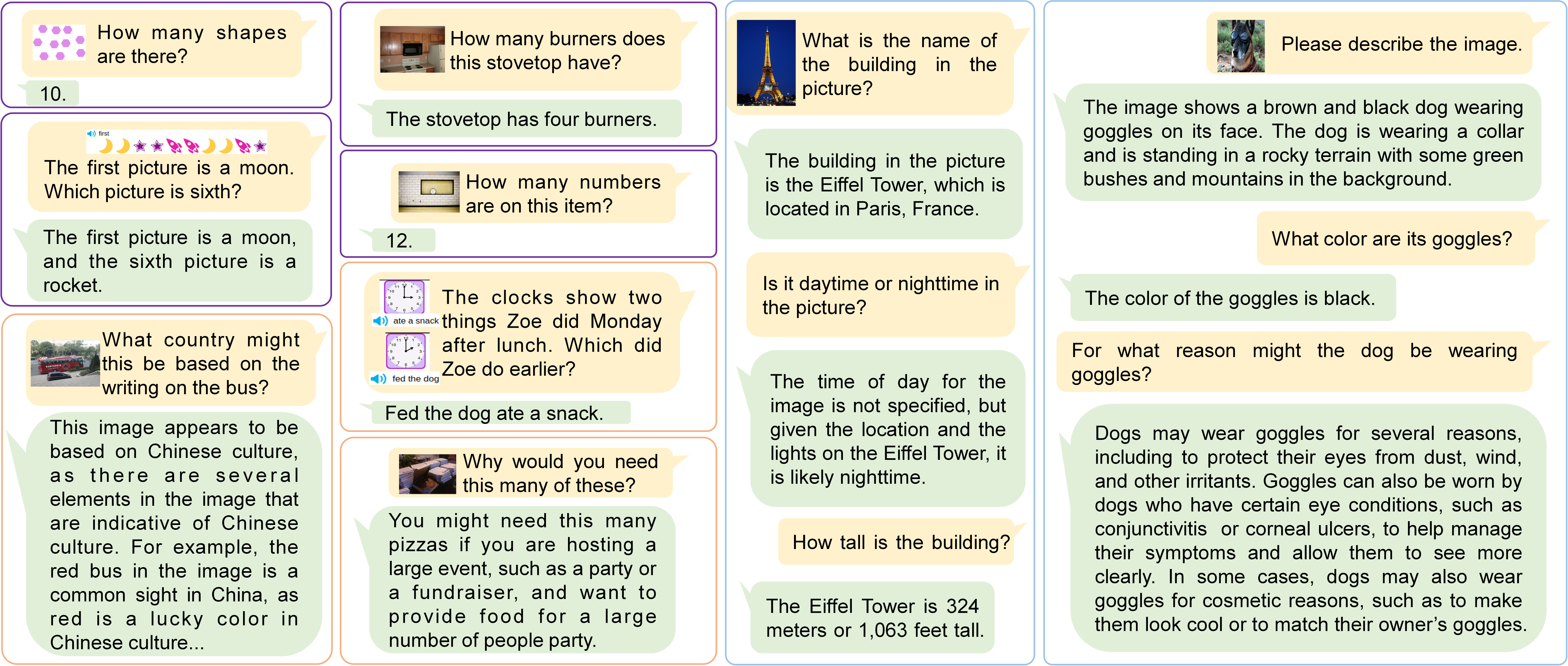}
\caption{\small{Examples of the test dialogs.}}
\label{app_con}
\end{figure*}

\section{Training Algorithm}
Note that we train the proposed VTC on ImageNet in an unsupervised manner. The training algorithm is summarized in algorithm.~\ref{alg}. We use FP16 in the model training. 
All experiments are conducted with a single Nvidia-A100(40G).

\begin{algorithm}[h]
\caption{{Training algorithm of VTC.}}
\label{alg}
\begin{algorithmic}
\STATE \textbf{Input}: Training image dataset $\mathcal{D}=\{\xv_i\}_{i=1}^N$, a pre-trained ViT $f$, VPG (\textit{e.g.}, Q-Former) and SD.\\
\STATE \textbf{Output}: Optimal VTC that completes the VPG by learning a set of reconstruction-aware tokens.
\STATE \textbf{Initialize}: All parameters in VTC.
\STATE \textbf{Preprocess}: Pre-calculate $\Wmat$ in Eq.(5) as described in~\cite{paischer2023sitta}.
\FOR{ iter = 1,2,3,...}
\STATE Sample a batch images from $\mathcal{D}$: $\xv^{(B)} = \{\xv_i\}_{i=1}^B$ and get their embeddings $\hv^{B}$.
\STATE \texttt{/*       Visual Token Complement     */}
\STATE Get the original vision tokens $\yv^{B}$ by feeding $\hv^{B}$ into the VPG.
\STATE Generate the reconstruction-aware visual tokens ${\yv'}^{B}$ with Eq.(3), and obtain the final completed prompt by concatenating $\yv^B$ with ${\yv'}^B$.
\STATE \texttt{/*       Diffusion-aware Loss     */}
\STATE Feed the visual prompt ${\hat{\yv}}^B$ into SD and compute the image reconstruction loss according to Eq.(4). 
\STATE Update all parameters in VTC with the stochastic gradient descent algorithm.
\ENDFOR\\
\end{algorithmic}\label{training}
\end{algorithm}

\section{Baselines}
To evaluate the performance of our approach, we compare VTC with recent state-of-the-art baselines:
\begin{itemize}
    \item Otter~\cite{li2023otter} is built on open-source version of Flamingo~\cite{awadalla2023openflamingo} and aligns the image and text under the in-context learning framework. We use the released code of Otter-7B to perform zero-shot generation.
    \item mPLUG-Owl~\cite{ye2023mplug} tunes the vision encoder and keeps LLMs fixed to align the vision and language embeddings in the first stage while further fine-tuning LLMs (with a LoRA module~\cite{hu2021lora}) and keeping the vision encoder fixed in the second instruction-tuning stage. We test its llama-7B version.
    \item VPGTrans~\cite{zhang2023transfer} aims to transfer an existing VPG from one MLLM to the target MLLM and develop a simple yet effective two-stage transfer framework, including a projector warm-up and vanilla fine-tuning. The test version is Vicuna-7B.
    \item LLaVA~\cite{liu2023llava} applies a single projection layer to convert image features from pre-trained CLIP visual encoder ViT-L/14 into the language embedding space of Vicuna-7B.
    \item BLIP2~\cite{blip2} is the first paper that designs the Q-Former as a VPG to bridge the image and text domains. It pre-trains Q-Former on web-scale text-caption pairs under the pre-training and fine-tuning pipelines. We use its FLAN-T5-XL version.
    \item MiniGPT-4~\cite{zhu2023minigpt} loads the pre-trained BLIP2 and aims to improve the alignment quality by fine-tuning a single projection layer in longer caption datasets. We load the Vincuna-7B version for \textit{LVLM-eHub} and \textit{DEMON} benchmarks and Vincuna-13B for \textit{MME}.
    \item InstructBLIP~\cite{zhang2023instruction} performs version-language instruction tuning based on the pre-trained BLIP2 by manually designing instruction templates for several vision tasks on collected multimodal datasets. We load the Vincuna-7B version for \textit{LVLM-eHub} and \textit{DEMON} benchmarks and FLAN-T5-XXL for \textit{MME}.
 \end{itemize}
 For all baselines, we report their results according to original papers or the corresponding benchmark leaderboards. We apply our proposed VTC on three well-known MLLMs (MiniGPT4, BLIP2, InstrctBLIP). To evaluate the robustness of VTC on base models with different structures and sizes, we load different versions of MiniGPT4 and InstructBLIP for different benchmarks. 
 This results in five versions of VTCs: VTC(BLIP2) with FLAN-T5-XL, VTC(MiniGPT4) with Vincuna-7B for \textit{LVLM-eHub} and \textit{DEMON} benchmarks, VTC(MiniGPT4) with Vincuna-13B for \textit{MME}, VTC(InstrcutBLIP) with Vincuna-7B for \textit{LVLM-eHub} and \textit{DEMON} benchmarks, and VTC(INstructBLIP) with FLAN-T5-XXL for \textit{MME}.

 \section{More Results and Analysis}
\renewcommand\thetable{\Alph{section}. \arabic{table}}   \renewcommand\thefigure{\Alph{section}. \arabic{figure}}    
\setcounter{table}{0}
\setcounter{figure}{0}

\textbf{Visualizations of the complementary tokens}. We provide more visualization results of the complementary tokens in Fig~\ref{app_park}. Thanks to the appended tokens, LLM captures more visual semantics and generates comprehensive image descriptions.

\textbf{More dialog examples}. We also provide extensive examples of the test dialogs in Fig~\ref{app_con}. We find that our VTC achieves impressive vision-language understanding in image captioning, object counting, and visual reasoning, across single and multi-turns.

 \section{ChatGPT 4(V) evaluation}
Here is the prompt used in the ChatGPT 4(v) evaluation: \\
\textit{``<IMG> Given the instruction \{instruction\}, please just provide the hallucination score[1-5] for the below answers without any explanation, where the fewer descriptive errors in the answer, the higher the hallucination score is given. The output format: Answer: score. Answer1: \{answer1\}, Answer2: \{answer2\},...''}.

\end{document}